\documentclass{article} 
\usepackage{hyperref}
\usepackage{url}
\usepackage{booktabs}
\usepackage{multirow}
\usepackage{amsfonts}
\usepackage{graphicx}
\usepackage{duckuments}
\usepackage{xcolor}
\usepackage{amsmath}
\usepackage{amssymb}
\usepackage{mathtools}
\usepackage{wrapfig}
\usepackage{tabularx}
\usepackage{svg}
\usepackage{array}
\usepackage{wrapfig}
\usepackage{adjustbox}
\usepackage{fontawesome5}
\usepackage{tcolorbox} 
\definecolor{cornflowerblue}{rgb}{0.39, 0.58, 0.93}
\hypersetup{
    colorlinks=true,
    linkcolor=cornflowerblue,
    filecolor=magenta,
    urlcolor=teal,
    citecolor=cornflowerblue,
    pdftitle={Spotting LLMs with Binoculars},
    pdfpagemode=FullScreen,
    }

\DeclareMathOperator{\PPL}{PPL} 
\DeclareMathOperator{\xPPL}{X-PPL} 

\newcommand{\prompt}{Write an article based on this ``Use an adjustable jump. Create your jump with cinder blocks and plywood. Use natural outdoor obstacles as jumps. Set down a treat on the far side of the jump. Practice at the lowest level until your dog is comfortable. Raise the bar gradually. Use the same method to teach your dog to jump through hoops."}

\newcommand{\defaultresponse}{A well-trained dog is not only a delight to be around, but it also provides a sense of pride and accomplishment to the owner. One captivating and useful skill that you may want to teach your dog is jumping. Whether you're doing it for fun, agility competitions, or simply to keep your furry companion physically engaged, teaching your dog how to jump can be a bonding and rewarding experience.}

\newcommand{\carlsagan}{Ah, my dear friend, we stand at the precipice of a most wondrous adventure, an odyssey of the canine kind. Our quest, to teach our loyal companions, those furry wonders of nature, to soar through the air with grace and agility. The tool at our disposal, the adjustable jump.}

\newcommand{\pirate}{Arrrr, me hearties! Ye landlubbers be wantin' yer pooch ta learn ta jump, eh? Well, fear not! I be here ta help ye teach yer furry first mate ta jump over obstacles like a proper salty sea dog! First things first, ye need ta create the jump. Now, I know what ye be thinkin': "Cinder blocks and plywood? That be a bit...rugged, don't ye think?" But never fear, me hearties!}

\newcommand{\nonrobotic}{Hey there, fellow dog lovers! Are you looking for a fun and challenging way to keep your furry friend active and engaged? Look no further than an adjustable jump! Not only is it a great way to provide your dog with physical exercise, but it's also a great way to bond with your pup.}

\newcommand{\update}[1]{{{#1}}}

\usepackage[accepted]{icml2024}

\icmltitlerunning{Spotting LLMs With Binoculars: Zero-Shot Detection of Machine-Generated Text}

\begin{document}

\twocolumn[
\icmltitle{Spotting LLMs With Binoculars: Zero-Shot Detection of ~~~~~~ Machine-Generated Text}
\icmlsetsymbol{equal}{*}

\begin{icmlauthorlist}
\icmlauthor{Abhimanyu Hans}{equal,umd}
\icmlauthor{Avi Schwarzschild}{equal,cmu}
\icmlauthor{Valeriia Cherepanova}{umd}
\icmlauthor{Hamid Kazemi}{umd}
\icmlauthor{Aniruddha Saha}{umd}
\icmlauthor{Micah Goldblum}{nyu}
\icmlauthor{Jonas Geiping}{ellis}
\icmlauthor{Tom Goldstein}{umd}
\end{icmlauthorlist}

\icmlaffiliation{umd}{University of Maryland}
\icmlaffiliation{cmu}{Carnegie Mellon University}
\icmlaffiliation{nyu}{New York University}
\icmlaffiliation{ellis}{ELLIS Institute Tübingen, MPI Intelligent Systems}

\icmlcorrespondingauthor{Abhimanyu Hans}{ahans1@umd.edu}
\icmlcorrespondingauthor{Avi Schwarzschild}{schwarzschild@cmu.edu}

\icmlkeywords{LLM, Machine Text Detection}

\vskip 0.3in]


\printAffiliationsAndNotice{\icmlEqualContribution} 

\begin{abstract}
Detecting text generated by modern large language models is thought to be hard, as both LLMs and humans can exhibit a wide range of complex behaviors. 
However, we find that a score based on contrasting two closely related language models is highly accurate at separating human-generated and machine-generated text.
Based on this mechanism, we propose a novel LLM detector that only requires simple calculations using a pair of pre-trained LLMs. 
The method, called {\em Binoculars}, achieves state-of-the-art accuracy without any training data.
It is capable of spotting machine text from a range of modern LLMs without any model-specific modifications.
We comprehensively evaluate {\em Binoculars} on a number of text sources and in varied situations. 
Over a wide range of document types, {\em Binoculars} detects over 90\% of generated samples from ChatGPT (and other LLMs) at a false positive rate of 0.01\%, despite not being trained on any ChatGPT data. 
Code available at \href{https://github.com/ahans30/Binoculars}{https://github.com/ahans30/Binoculars}.
\end{abstract}

\begin{figure*}[t!]
    \centering
    \includegraphics[width=0.75\textwidth]{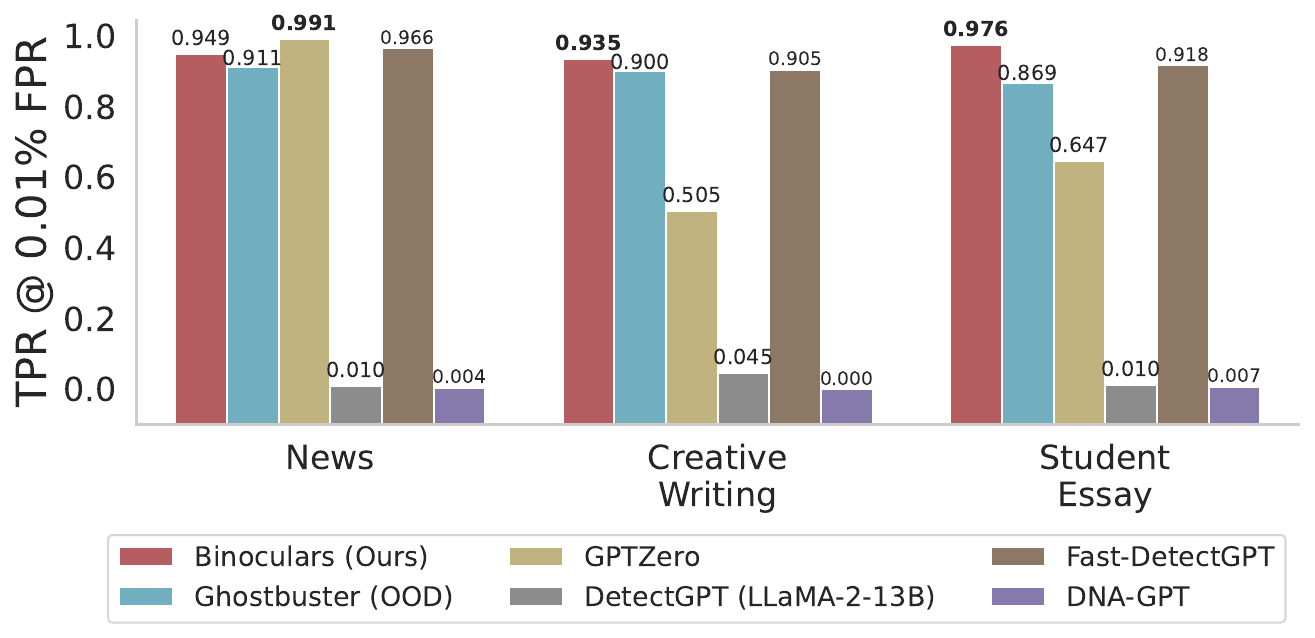}
    \caption{\textbf{Detection of Machine-Generated Text from ChatGPT}. Our detection approach using \textit{Binoculars} is highly accurate at separating machine-generated and human-written samples from \textit{News}, \textit{Creative Writing} and \textit{Student Essay} datasets with a false positive rate of $0.01\%$. \textit{Binoculars}, based on open-source Falcon models with no finetuning, outperforms commercial detection systems, such as GPTZero, as well as open-source detectors -- even though both of these baselines are specifically tuned to detect ChatGPT \citep{verma_ghostbuster_2023,tian_gptzero_2023}. Our approach operates entirely in a zero-shot setting and has not been tuned on ChatGPT specifically.}
    \label{fig:performance}
\end{figure*}

\section{Introduction}
\label{sec:intro}

We present a method to detect LLM-generated text that works in the zero-shot setting in which no training examples are used from the LLM source.  Even with this strict limitation, our scheme still out-performs all open-source methods for ChatGPT detection and is competitive with or better than commercial APIs, despite these competitors using training samples from ChatGPT~\citep{mitchell_detectgpt_2023,verma_ghostbuster_2023}.
At the same time, because of the zero-shot nature of our detector, the very same detector can spot multiple different LLMs with high accuracy---something that all existing solutions fail to do.

The ability to detect LLMs in the zero-shot setting addresses issues of growing importance.  
Prior research on combating academic plagiarism \citep{turnitin} has fixated strongly on ChatGPT because of its simple and accessible interface. 
But more sophisticated actors use LLM APIs to operate bots, create fake product reviews, and spread misinformation on social media platforms at a large scale. 
These actors have a wide range of LLMs available to them beyond just ChatGPT, making zero-shot, model-agnostic detection critical for social media moderation and platform integrity assurance~\citep{crothers_machine_2022, bail_difficulty_2023}.  
Our zero-shot capability is a departure from existing detectors that rely on model-specific training data and often fail to transfer to new models. 


Our proposed detector, called {\em Binoculars}, works by viewing text through two lenses. 
First, we compute the $\log$ perplexity of the text in question using an ``observer'' LLM.
We then compute next-token predictions using a ``performer'' LLM and compute their perplexity according to the observer. We call this metric \textit{cross-perplexity}. We observe that perplexity per cross-perplexity is a surprisingly powerful signal to detect LLM-text.
We first motivate this simple observation, and then show that it is sufficient to build a strong zero-shot detector, which we extensively stress-test in a number of text domains.


\section{The LLM Detection Landscape}
\label{sec:related-work}


Successful efforts to spot machine-generated text show promise on early models whose generation output is not convincingly human.
However, with the rise of transformer models for language modeling \citep{radford_language_2019,brown_language_2020,chowdhery_palm_2022,touvron_llama_2023-1}, primitive mechanisms to detect machine-generated text are rendered useless.
While one approach is to record \citep{krishna_paraphrasing_2023} or watermark all generated text \citep{kirchenbauer_watermark_2023}, these \emph{preemptive detection} approaches can only be implemented with full control over the generative model. 

Instead, the recent spread of machine-generated text, especially via ChatGPT, has led to a flurry of work on \textit{post-hoc detection} approaches that can be used to detect machine text without cooperation from model owners. 
These detectors can be separated into two main groups. 
The first is trained detection models, where a pretrained language model backbone is finetuned for the binary classification task of detection \citep{solaiman_release_2019,zellers_defending_2019,yu_gpt_2023,zhan_g3detector_2023}, including techniques like adversarial training \citep{hu_radar_2023} and abstention \citep{tian_multiscale_2023}. 
Alternatively, instead of finetuning the whole backbone, a linear classifier can be fit on top of frozen learned features, which allows for the inclusion of commercial API outputs  \citep{verma_ghostbuster_2023}.

The second category of approaches comprises statistical signatures that are characteristic of machine-generated text. 
These approaches have the advantage of requiring none or little training data and they can easily be adapted to newer model families \citep{pu_unraveling_2022}. 
Examples include detectors based on perplexity \citep{tian_new_2023, vasilatos_howkgpt_2023,wang_m4_2023}, perplexity curvature \citep{mitchell_detectgpt_2023}, log rank \citep{su_detectllm_2023}, intrinsic dimensionality of generated text \citep{tulchinskii_intrinsic_2023}, and n-gram analysis \citep{yang_dna-gpt_2023}. 
Our coverage of the landscape is non-exhaustive, and we refer to recent surveys \citet{ghosal2023towards,tang_science_2023,dhaini_detecting_2023,guo_how_2023}  for additional details. 

From a theoretical perspective, \citet{varshney_limits_2020}, \citet{helm_statistical_2023}, and \citet{sadasivan_can_2023} all discuss the limits of detection.
These works generally agree that fully general-purpose models of language would be, by definition, impossible to detect. 
However, \citet{chakraborty_possibilities_2023} note that even models that are arbitrarily close to this optimum are technically detectable given a sufficient number of samples. 
In practice, the relative success of detection approaches, such as the one we propose and analyze in this work, provides constructive evidence that current language models are imperfect representations of human writing -- and thereby detectable. 
Finally, the robustness of detectors to attacks attempting to circumvent detection can provide stronger practical limits on reliability in the worst case \citep{bhat_how_2020,wolff_attacking_2022,liyanage_detecting_2023}.

With an understanding of how much work exists on LLM detection, a crucial question arises: How do we appropriately and thoroughly evaluate detectors? 
Many works focus on accuracy on balanced test sets and/or AUC of their proposed classifiers, but these metrics are not well-suited for the high-stakes question of detection. 
Ultimately, only detectors with low false positive rates across a wide distribution of human-written text, truly reduce harm. 
Further, \citet{liang_gpt_2023} note that detectors are often only evaluated on relatively easy datasets that are reflective of their training data. 
Their performance on out-of-domain samples is often abysmal.
For example, TOEFL essays written by non-native English speakers were wrongly marked as machine-generated 48-76\% of the time by commercial detectors \citep{liang_gpt_2023}.

In Section~\ref{sec:methods}, we motivate our approach and discuss why detecting language model text, especially in the ChatGPT world, is difficult.
In this work, our emphasis is directed toward baselines that function within post-hoc, out-of-domain (zero-shot), and black-box detection scenarios. We use the state-of-the-art open source detector Ghostbuster \citep{verma_ghostbuster_2023}, the commercially deployed GPTZero\footnote{\href{https://gptzero.me/}{https://gptzero.me/}}, DetectGPT \citep{mitchell_detectgpt_2023} and its efficient version Fast-DetectGPT \citep{fastdetectgpt} and finally DNA-GPT \citep{dnagpt2023} to compare detection performance across various datasets in Section~\ref{sec:results}. In Section~\ref{sec:edge-cases}, we evaluate the reliability of \emph{Binoculars} in various settings that constitute edge cases and crucial deployment behaviors that a detector based on \emph{Binoculars} has to take into account. Please see appendix \ref{baselines-impl-deets} for more details.

\section{\emph{Binoculars}: How it works}
\label{sec:methods}

Our approach, \emph{Binoculars}, is so named as we look at inputs through the lenses of two different language models.
It is well known that perplexity -- a common baseline for machine/human classification -- is insufficient on its own, leading prior work to disfavor approaches based on statistical signatures. 
However, we propose using a ratio of perplexity measurement and \textit{cross-perplexity}, a notion of how surprising the next token predictions of one model are to another model. 
This two-model mechanism is the basis for our general and accurate detector, and we show that this mechanism is able to detect a number of LLMs, even when they are unrelated to the two models used by \emph{Binoculars}.

\subsection{Background \& Notation}
\label{sec:background}

A string of characters $s$ can be parsed into tokens and represented as a list of token indices $\vec x$ by a tokenizer $T$. 
Let $x_i$ denote the token ID of the $i$-th token, which refers to an entry in the LLMs vocabulary $V = \{1, 2..., n\}$.
Given a token sequence as input, a language model $\mathcal{M}$ predicts the next token by outputting a probability distribution over the vocabulary: 

\begin{equation}
    \begin{aligned} \ignorespacesafterend
    &\mathcal M(T(s)) = \mathcal M(\vec x) = Y \\
    &Y_{ij} = P(v_j | x_{0:i-1})  \text{ for all } j \in V.   
    \end{aligned}
\end{equation}

We abbreviate $\mathcal M(T(s))$ as $\mathcal M(s)$ where the tokenizer is implicitly the one used in training $\mathcal M$.
We define $\log \PPL$, the log-perplexity, as the average negative log-likelihood of all tokens in the given sequence.
Formally, let
\begin{equation}
    \begin{aligned}
    &\log \PPL_\mathcal M (s) = - \frac{1}{L} \sum_{i = 1}^L \log(Y_{ix_i}), \\
    &\text{ where }\vec x = T(s),~ Y = \mathcal M(\vec x), \\&\text{ and }
    L = \text{ number of tokens in }s.
    \end{aligned}
\end{equation}
Intuitively, log-perplexity measures how ``surprising'' a string is to a language model. As mentioned above, perplexity has been used to detect LLMs, as humans produce more surprising text than LLMs.
This is reasonable, as $\log \PPL$ is also the loss function used to train generative LLMs, and models are likely to score their own outputs as unsurprising.
Our method also measures how surprising the output of one model is to another.
We define the \emph{cross-perplexity}, which takes two models and a string as its arguments.
Let $\update{\log} \xPPL_{\mathcal M_1, \mathcal M_2}(s)$ measure the average per-token cross-entropy between the outputs of two models, $\mathcal M_1$ and $\mathcal M_2$ , when operating on the tokenization of $s$.\footnote{This requires that $\mathcal M_1$ and $\mathcal M_2$ share a tokenizer.}
\begin{equation}
    \update{\log} \xPPL_{\mathcal M_1, \mathcal M_2}(s) = - \frac{1}{L} \sum_{i = 1}^L \mathcal M_1(\update{s})_i \cdot \log\left(\mathcal M_2(\update{s})_i\right).
    \label{cross-ppl-equation}
\end{equation}
Note that $\cdot$ denotes the dot product between two vector-valued quantities.

\subsection{What makes detection hard? A primer on the capybara problem.}

Why do we require measurements of both perplexity and cross-perplexity? Unsurprisingly, LLMs tend to generate text that is unsurprising to an LLM. Meanwhile, because humans differ from machines, human text has higher perplexity according to an LLM observer. For this reason, it is tempting to use raw perplexity for LLM detection, as high perplexity is a strong sign of a human author.

Unfortunately, this intuition breaks when hand-crafted prompts are involved. Prompts have a strong influence over downstream text, and prompts are typically unknown to the detector.  On the one hand, the prompt ``1, 2, 3,'' might result in the very low perplexity completion ``4, 5, 6.''  On the other hand, the prompt  ``Can you write a few sentences about a capybara that is an astrophysicist?'' will yield a response that seems more surprising. In the presence of a prompt, the response may be unsurprising (low perplexity). 
 But in the absence of the prompt, a response containing the curious words ``capybara'' and ``astrophysicist'' in the same sentence will have high perplexity, resulting in the false determination that the text was written by a human, see the example in Table \ref{tab:capybara}.  Clearly, certain contexts will result in high perplexity and others low perplexity, regardless of whether the author is human or machine. We refer to this dilemma as ``the capybara problem'' -- in the absence of the prompt, LLM detection seems difficult and naive perplexity-based detection fails. 

\begin{table*}[ht!]
    \centering
     \begin{tcolorbox}[left=0mm,right=0mm,width=\textwidth,bottom=0mm,top=0mm,arc=0mm,auto outer arc]
    \begin{tabular}{m{0.1in}m{1.5in}m{1.5in}m{1.5in}}
         &\multicolumn{3}{p{5.9in}}{\looseness -1``Dr. Capy Cosmos, a capybara unlike any other, astounded the scientific community with his groundbreaking research in astrophysics. With his keen sense of observation and unparalleled ability to interpret cosmic data, he uncovered new insights into the mysteries of black holes and the origins of the universe. As he peered through telescopes with his large, round eyes, fellow researchers often remarked that it seemed as if the stars themselves whispered their secrets directly to him. Dr. Cosmos not only became a beacon of inspiration to aspiring scientists but also proved that intellect and innovation can be found in the most unexpected of creatures.'' -- GPT 4}\\
    \end{tabular}
    \caption{This quote is LLM output from ChatGPT (GPT-4) when prompted with ``Can you write a few sentences about a capybara that is an astrophysicist?'' The Falcon LLM assigns this sample a high perplexity (2.20), well above the mean for both human and machine data. Despite this problem, our detector correctly assigns a \emph{Binoculars} score of 0.73, which is well below the global threshold of 0.901, resulting in a correct classification with high confidence. For reference, DetectGPT wrongly assigns a score of 0.14, which is below its threshold of 0.17, and classifies the text as human. GPTZero assigns a 49.71\% score that this text is generated by AI. }
    \label{tab:capybara}
     \end{tcolorbox}
\end{table*}

\subsection{Our Detection Score}
\label{sec:our-score}
 
{\em Binoculars} solves the \textit{capybara problem} by providing a mechanism for estimating the baseline perplexity induced by the prompt. By comparing the perplexity of the observed text to this expected baseline, we observe a surprising LLM text signature.


\update{
\textbf{Motivation.} 
LLM-generated text may exhibit a high perplexity score depending on the prompt specified which yields a simple perplexity-based detector ineffective (see the ``Capybara Problem" in Table~\ref{tab:capybara}). To calibrate for prompts that yield high-perplexity generation, we use \emph{cross-perplexity} introduced Equation \eqref{cross-ppl-equation} as a normalizing factor that intuitively encodes the perplexity level of next-token predictions from two models.
}

Rather than examining raw perplexity scores, we instead propose measuring whether the tokens that appear in a string are surprising {\em relative to the baseline perplexity of an LLM acting on the same string.}  
A string might have properties that result in high perplexity when completed by any agent, machine or human.  
Yet, we expect the next-token choices of humans to be even higher perplexity than those of a machine.  
By normalizing the observed perplexity by the expected perplexity of a machine acting on the same text, we can arrive at a detection metric that is fairly invariant to the prompt; see Table~\ref{tab:capybara}.

We propose the {\em Binoculars} score $B$ as a sort of normalization or reorientation of perplexity.
In particular, we look at the ratio of perplexity to cross-perplexity. 
\begin{equation}\label{eq:binoculars}
    B_{\mathcal M_1, \mathcal M_2}(s) = \frac{\update{\log}\PPL_{\mathcal M_1}(s)}{\update{\log}\xPPL_{\mathcal M_1, \mathcal M_2}(s)}
\end{equation}

Here, the numerator is simply the perplexity, which measures how surprising a string is to $\mathcal M_1$.
The denominator measures how surprising the token predictions of $\mathcal M_2$ are when observed by $\mathcal M_1.$  Intuitively, we expect a human to diverge from  $\mathcal M_1$ more than  $\mathcal M_2$ diverges from  $\mathcal M_1$, provided the LLMs $\mathcal M_1$ and  $\mathcal M_2$ are more similar to each other than they are to a human.  

The \emph{Binoculars} score is a general mechanism that captures a statistical signature of machine text. In the sections below, we show that for most obvious choices of $\mathcal M_1$ and $\mathcal M_2$, \emph{Binoculars} separates machine and human text better than perplexity alone---and it is capable of detecting generic machine text generated by a third model altogether. 

\looseness -1 Interestingly, we can draw some connection to other approaches that contrast two language models, such as contrastive decoding \citep{li_contrastive_2023}, which aims to generate high-quality text completions by generating text that roughly maximizes the difference between a weak and a strong model.
Speculative decoding is similar~\citep{chen_accelerating_2023,leviathan_fast_2023}, it uses a weaker model to plan completions. 
Both approaches function best when pairing a strong model with a very weak secondary model. 
However, as we show below, our approach works best for two models that are very close to each other in performance. In the remainder of this work, to determine Binoculars score in the Equation \ref{eq:binoculars} we compute the numerator i.e. $\PPL$ using the Falcon-7B-Instruct model \citep{almazrouei_falcon-40b_2023} and compute denominator i.e. $\xPPL$ with Falcon-7B and Falcon-7B-Instruct respectively. The full set of combinations of scoring models used can be found in Table~\ref{ablation:other-scorers} in the appendix with the first row depicting the default formulation used through main paper body.

\section{Accurate Zero-Shot Detection}
\label{sec:results}

In this section, we evaluate \emph{Binoculars} as a zero-shot LLM detector in multiple domains. In our experiments, we focus on the problem setting of detecting machine-generated text from a modern LLM, as generated in common use cases without consideration for the detection mechanism. 

\begin{figure*}[t!]
    \centering
    \includegraphics[width=0.8\textwidth]{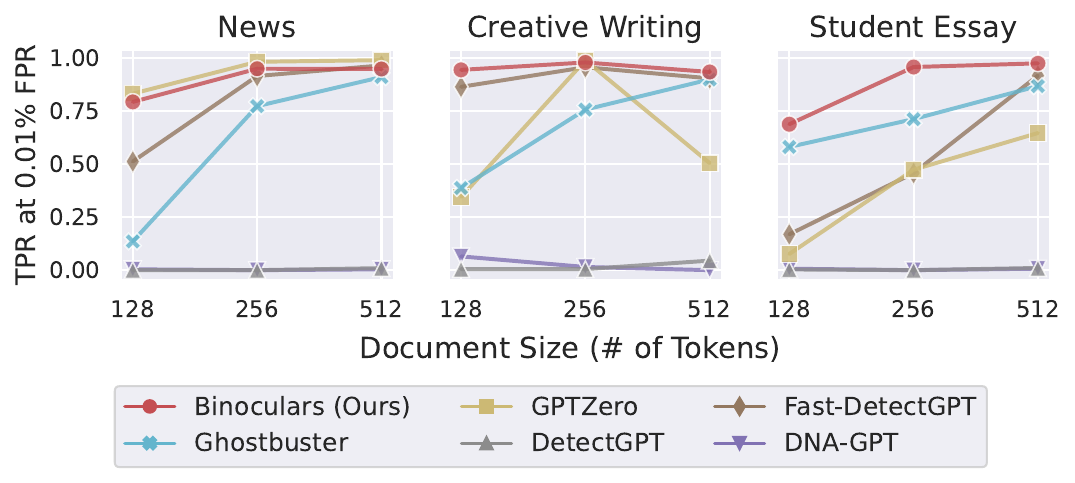}
    \caption{\textbf{Impact of Document Size on Detection Performance}. The plot displays the TPR at 0.01\% FPR across varying document sizes by prefixing sample documents. The x-axis represents the number of tokens of the observed document, while the y-axis indicates the corresponding detection performance, highlighting the \emph{Binoculars} ability to detect with a low number of tokens.}
\label{fig:perf-ours-vs-gh}
\end{figure*}

\subsection{Datasets}
\label{sec:data}

We start our experiments with several datasets described in the LLM detection literature. 
The most recent baseline to which we compare is Ghostbuster. 
\citet{verma_ghostbuster_2023}, who propose this method, also introduce three datasets that we include in our study: \textit{Writing Prompts}, \textit{News}, and \textit{Student Essay}.
These are balanced datasets with equal numbers of human samples and machine samples. The machine samples are written by ChatGPT.\footnote{In the following we will always use ChatGPT as short-hand for the chat versions of GPT-3.5-(turbo).}

We also generate several datasets of our own to evaluate our capability in detecting other language models aside from ChatGPT. Drawing samples of human-written text from CCNews \citep{Hamborg2017}, PubMed \citep{sen_collective_2008}, and {CNN} \citep{hermann_teaching_2015}, we generate corresponding, machine-generated completions using LLaMA-2-7B and Falcon-7B (see details in Appendix \ref{sec:appendix-experiments-data-gen}). 
Further, we use the Orca dataset \citep{OpenOrca}, which provides several million instruction prompts with their machine-generated completions from GPT-3 and GPT-4. 


\subsection{Metrics}
\label{sec:metrics}

Since detectors are binary classifiers, the standard suite of binary classification metrics is relevant. It is often considered comprehensive to look at ROC curves and only report under the curve (AUC) as a performance metric.
In fact, \citet{verma_ghostbuster_2023} and \citet{mitchell_detectgpt_2023} only report performance as measured by AUC and F1 scores. We argue that these metrics alone are inadequate when evaluating LLM detection performance.


In LLM detection, the most concerning harms often arise from {\em false positives}, i.e., instances when human text is wrongly labeled as machine-generated. 
For this reason, we focus on true-positive rates (TPR) at low false-positive rates (FPR), and adopt a standard FPR threshold of $0.01\%$.\footnote{The smallest threshold we can comprehensively evaluate to sufficient statistical significance with our compute resources.}
We also note that AUC scores are often uncorrelated with TRP$@$FPR when the FPR is below $1\%$ (Table \ref{tab:ppl-v-xppl-results}). 
When evaluating F1-Score, we purely use the ``out-of-domain'' threshold. (see Appendix \ref{sec:appendix-threshold} for details).

\subsection{Benchmark Performance}
\label{sec:benchmarks}

Using a handful of datasets, we compare the AUC and TPR of \emph{Binoculars} to Ghostbuster \citep{verma_ghostbuster_2023}, GPTZero \citep{tian_gptzero_2023}, the commercially deployed
GPTZero, and DetectGPT (using LLaMA-2-13B to score curvature) \citep{mitchell_detectgpt_2023}. 
We highlight that these comparisons on machine samples from ChatGPT are \textit{in favor} of GPTZero and Ghostbuster, as these detectors have been tuned to detect ChatGPT output, and comparisons using samples from LLaMA models are \textit{in favor} of DetectGPT for the same reason.

\textbf{Ghostbuster Datasets.} 
The Ghostbuster detector is a recent detector tuned to detect output from ChatGPT.
Using the same three datasets introduced and examined in the original work by \citet{verma_ghostbuster_2023}, we compare TPR at 0.01\% FPR in Figure~\ref{fig:performance} (and F1-Score in Figure~\ref{fig:gh-dataset-performance} in Appendix) to show that \textit{Binoculars} outperforms  Ghostbuster in the ``out-of-domain'' setting. 
A desirable property for detectors is that with more information they get stronger. Figure~\ref{fig:perf-ours-vs-gh} shows that both \emph{Binoculars} and Ghostbuster have this property, and that the advantages of \textit{Binoculars} are even clearer in the few-token regime.


\textbf{Open-Source Language Models.}
We show that our detector is capable of detecting the output of several LLMs, such as LLaMA as shown in Figure~\ref{fig:core-perf-ours-vs-gh-roc} and Falcon as shown in Figure~\ref{fig:performance_our_datasets} in the appendix. 
Here we also observe that Ghostbuster is indeed only capable of detecting ChatGPT, and it fails to reliably detect LLaMA generated text. The detailed ROC plots in Figure~\ref{fig:core-perf-ours-vs-gh-roc} compare performance across thresholds for all methods.

\begin{figure*}[tb!]
    \centering
    \includegraphics[width=0.8\textwidth]{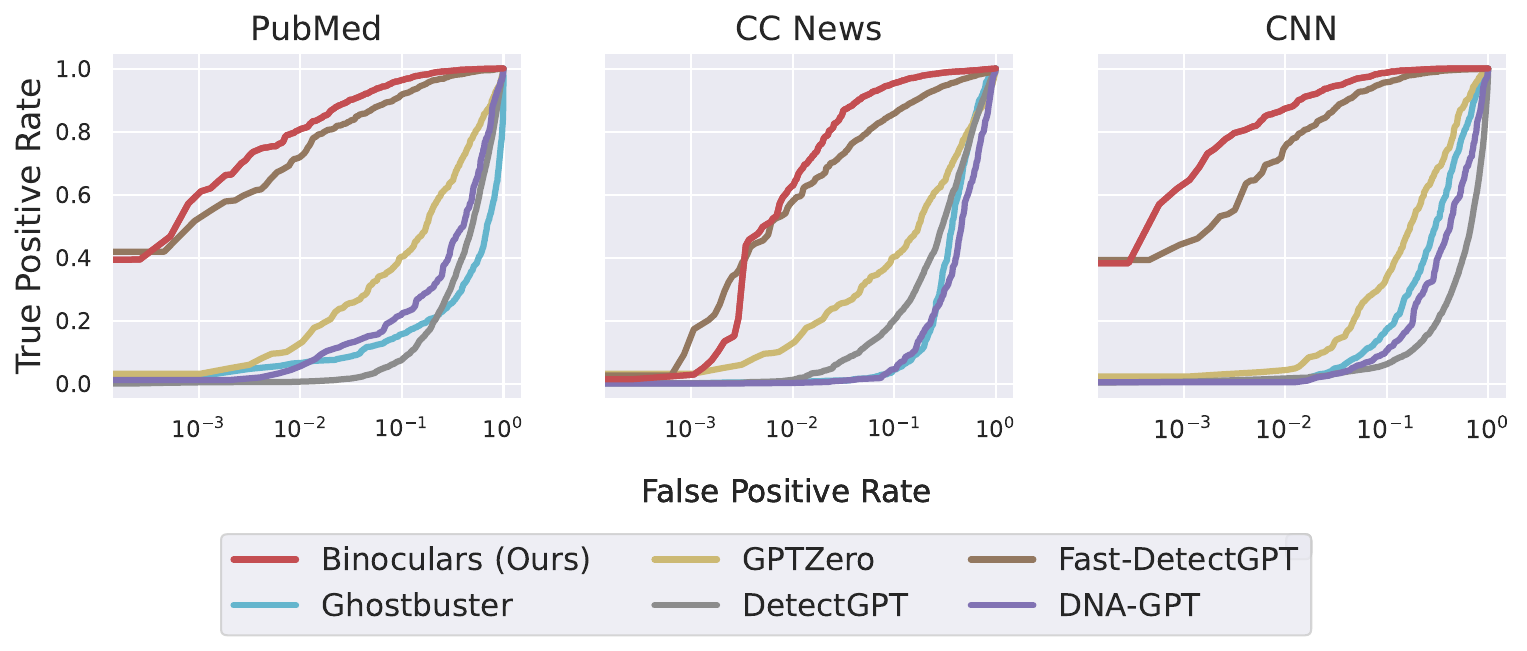}
    \caption{\textbf{Detecting LLaMA-2-13B generations.} \emph{Binoculars} achieves higher TPRs for low FPRs (on log scale) than other methods.}
    \label{fig:core-perf-ours-vs-gh-roc}
\end{figure*}

\section{Reliability in the Wild}
\label{sec:edge-cases}

How well does \textit{Binoculars} work when faced with scenarios encountered in the wild? 
The key takeaway that we want to highlight throughout this section is that the score underlying \textit{Binoculars}, i.e. Equation~\eqref{eq:binoculars} is a \emph{machine-text detector}. Intuitively, this means that is predicts how likely it is that the given piece of text could have been generated by a similar language model. 
This has a number of crucial implications regarding memorized samples, text from non-native speakers, modified prompting strategies, and edge cases, all of which we comprehensively evaluate in this section. 

\begin{figure*}[t!]
    \centering
    \includegraphics[trim=0mm 2mm 0mm 2mm,clip,width=0.9\textwidth]{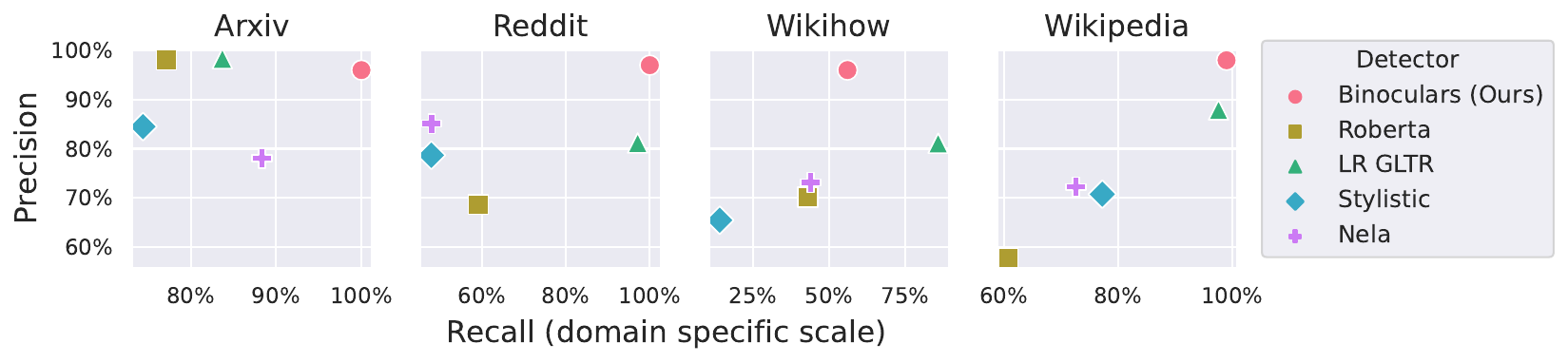}
    \caption{\update{\textbf{Detection of ChatGPT-generated text in various domains from M4 Dataset.} Binoculars is more precise over 4 domains using the OOD threshold for detection. We use the mean of out-of-domain performance metrics reported by \citet{wang_m4_2023}}}
    \label{fig:m4-domain-with-baselines}
\end{figure*}

\vspace{-2pt}

\subsection{Varied Text Sources}

\begin{figure}[tb!]
    \centering
    \includegraphics[trim=0mm 2mm 0mm 2.5mm,clip,width=0.8\columnwidth]{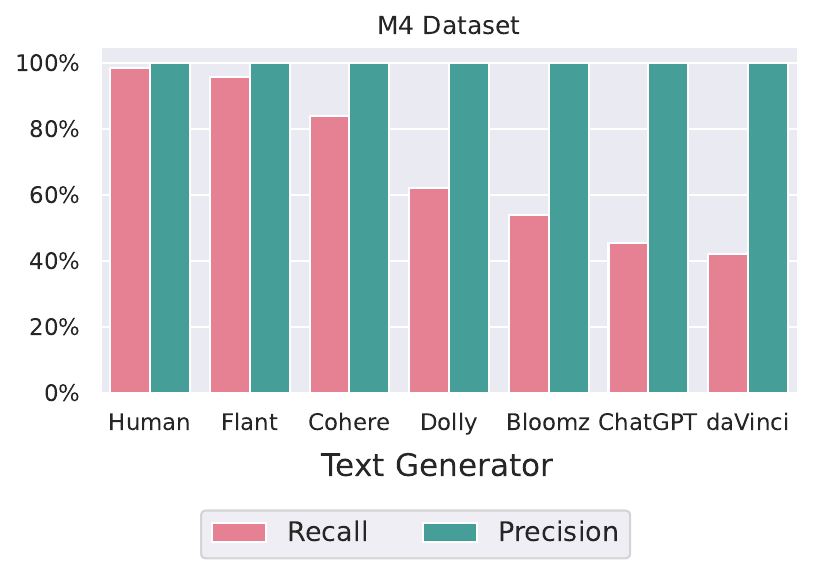}
    \caption{Performance of \emph{Binoculars} on samples \\from various generative models.}
    \label{fig:m4}
\end{figure}

We start our investigation by exploring detector performance in additional settings outside of the English language. To this end we investigate the Multi-generator, Multi-domain, and Multi-lingual (M4) detection datasets \citep{wang_m4_2023}.
These samples come from Arxiv, Reddit, Wikihow, and Wikipedia sources, and include examples in varied languages, such as Urdu, Russian, Bulgarian, and Arabic. 
Machine text samples in this dataset are generated via ChatGPT.
In Figure~\ref{fig:m4-domain-with-baselines}, we show the precision and recall of \emph{Binoculars} and four other baselines, showing that our method generalizes across domains and languages. 
These baselines, released with the M4 Datasets, include fine-tuned RoBERTa \citep{liu2019roberta}  classifier ~\citep{zellers_defending_2019, solaiman_release_2019}, Logistic Regression over Giant Language Model Test Room (LR GLTR) \citep{gltr-baseline} which generates features assuming predictions are sampled from a specific token distribution, Stylistic \citep{stylistic-baseline} which employs syntactic features at character, word, and sentence level, News Landscape classifiers (NELA) \citep{nela-baseline} which generates and leverages semantic and structural features for veracity classification. 
We reprint this result from the benchmark for reference. Results with more source models appear in Figure~\ref{fig:m4}.


\begin{figure}[tb!]
    \centering
    \includegraphics[trim=0mm 3mm 0mm 0mm,clip,width=.6\columnwidth]{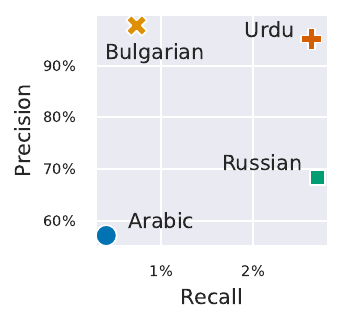}
    \caption{\emph{Binoculars} operates at high precision in Bulgarian and Urdu, but with low recall in all four languages.} 
    \label{fig:multi-lingual}
\end{figure}

\subsection{Other languages}
\label{sec:other-languages}

When evaluating \emph{Binoculars} on languages that are not well represented in Common Crawl data (standard LLM pretraining data), we find that false-positive rates remain low, which is highly desirable from a harm reduction perspective.
However, machine text in these low-resource languages is often classified as human. Figure~\ref{fig:multi-lingual} shows that we indeed have reasonable precision but poor recall in these settings. While this ordering of scores is a win for harmlessness, why is multilingual text detection limited?


Due to the limited capability of Falcon models (powering \emph{Binoculars}) in generating text in these low-resource languages in our experiments, we hypothesize that a stronger multilingual pair of models would lead to a version of \textit{Binoculars} that could spot ChatGPT-generated text in these languages more effectively.

\vspace{-5pt}

\begin{figure}[t!]
    \centering
    \centering
    \includegraphics[width=0.8\columnwidth]{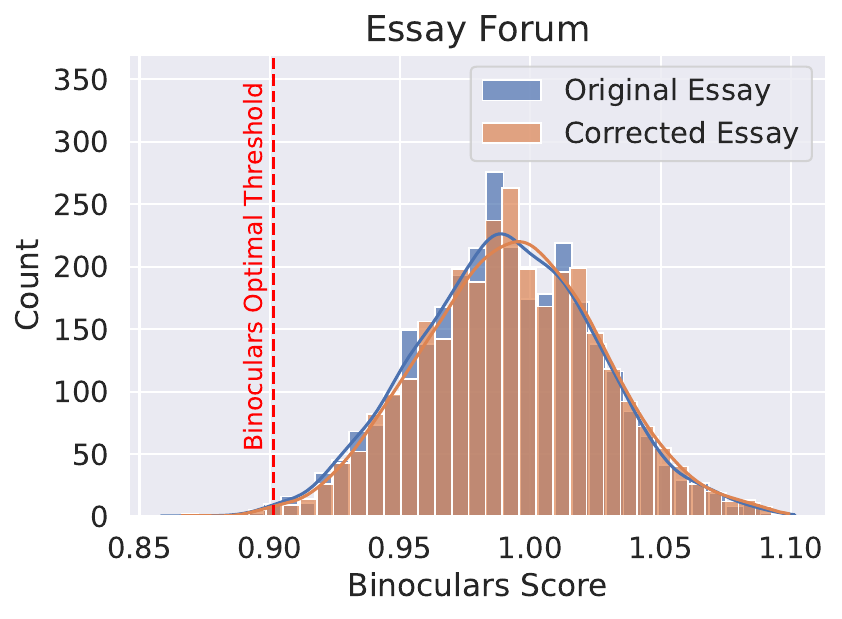}
    \caption{The distribution of \emph{Binoculars} scores remains unchanged when the English grammar is corrected in essays composed by non-native speakers. Both original and corrected essays are unambiguously classified as human-written.} 
    \label{fig:essay-forum}
\end{figure}

\paragraph{False-positive rates on non-native speakers' writing} 
A significant concern about LLM detection algorithms, as raised by \citet{liang_gpt_2023}, is that LLM detectors are inadvertently biased against non-native English speakers (ESL) classifying their writing as machine-generated exceedingly often.
To test this, we analyze essays from \textit{EssayForum}, a web page for ESL students to improve their academic writing~\citep{essayforum_nid989essayfroum-dataset_2022}.
This dataset contains both the original essays, as well as grammar-corrected versions.
We compare the distribution of \emph{Binoculars} scores across the original and the grammar-corrected samples.
Interestingly, and in stark comparison to commercial detectors examined by \citet{liang_gpt_2023} on a similar dataset, \emph{Binoculars} attains equal accuracy at 99.67\% for both corrected and uncorrected essay datasets (see Figure~\ref{fig:essay-forum}). 
We also point out that the \emph{Binoculars} score distribution on ESL's text highly overlaps with that of grammar-corrected versions of the same essays, showing that detection through \textit{Binoculars} is insensitive to this type of shift. 

\subsection{Memorization}
\label{sec:memorized-strings}

One common feature of perplexity-based detection is that highly memorized examples are classified as machine-generated.
For example, famous quotes that appear many times in the training data likely have low perplexity according to an observer model that has overfit to these strings. 
By looking at several examples, we examine how \emph{Binoculars} performs on this type of data.

See Table~\ref{famous-quotes} in Appendix~\ref{sec:app-famous-text} for all famous texts evaluated in this study.
First, we ask about the US Constitution -- a document that is largely memorized by modern LLMs. 
This example has a \emph{Binoculars} score of 0.76, well into the machine range. Of the 11 famous texts we study, this was the lowest score (most \emph{machine-y}). Three of the 11 fall on the machine-side of our threshold.
It is important to note that while this behavior may be surprising, and does require careful consideration in deployment, it is fully consistent with a machine-text detector.
Memorized text is both text written by human writers, and text that is likely to be generated by an LLM.  
Classification of memorized text as machine generated may be acceptable or even desirable in some applications (e.g., plagiarism detection), or undesirable in others (e.g., removal of LLM-generated text from a training corpus).


\begin{figure}[tb!]
    \centering
    \includegraphics[trim=0mm 0mm 0mm 2mm,clip,width=\columnwidth]{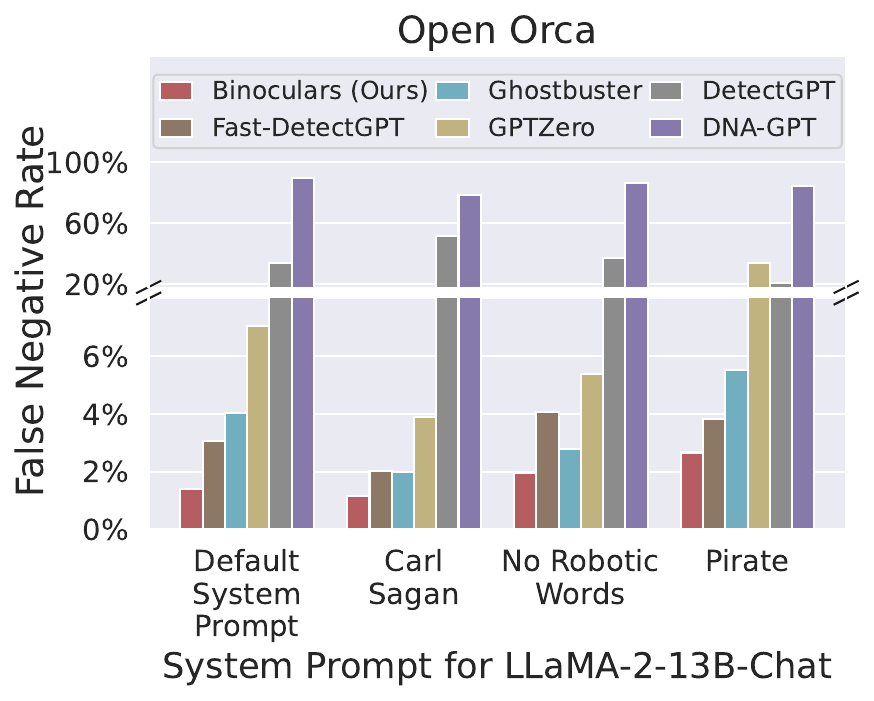}
\caption{Detection with modified system prompts.} 
\label{fig:m4-prompting-strategy}
\end{figure}

\begin{table*}[t!]
\begin{center}
\footnotesize
\caption{\footnotesize Excerpts from LLaMA-2-13B-chat generations using an Open-Orca sample prompt with varying system prompts. For the full modifications to the prompts to generate these stylized outputs, see Appendix~\ref{sec:app-modified-prompts}.}
\label{tab:modified-prompts}
\begin{tabular}{  >{\arraybackslash}p{0.13\linewidth}  m{5.3in}  } 
 \toprule
 Prompt & \scriptsize{\prompt} \\ 
 \hline
 \hline
 Default & \scriptsize{\defaultresponse} \\ 
 \hline
 Carl Sagan & \scriptsize{\carlsagan} \\ 
 \hline
 Non-Robotic & \scriptsize{\nonrobotic} \\ 
 \hline
 Pirate & \scriptsize{\pirate} \\
 \bottomrule
\end{tabular}
\label{tab:orca-samples}
\end{center}
\end{table*}

\subsection{Modified Prompting Strategies}
\label{sec:modified-prompts}
 
The Open Orca dataset contains machine generations from both GPT-3 and GPT-4 for a wide range of tasks \citep{OpenOrca}.
This serves as a diverse test bed for measuring \emph{Binoculars} on both of these modern high-performing LLMs.
Impressively, \emph{Binoculars} detects 92\% of GPT-3 samples and 89.57\% of GPT-4 samples when using the global threshold (from reference datasets). Note, we only report accuracy since this is over a set of machine-generated text only.
This dataset also allows us to explore how sensitive \emph{Binoculars} is to modifying prompts.

Simple detection schemes are sometimes fooled by simple changes to prompting strategies, which can produce stylized text that deviates from the standard output distribution.
With this in mind, we use LLaMA-2-13B-chat and prompts designed to tweak the style of the output.
Specifically, we prompt LLaMA2-13B-chat with three different system prompts by appending to the standard system prompt a request to write in Carl Sagan's voice or without any mechanical or robotic sounding words or like a pirate. 

In general, we find that these stylistic changes do not significantly impact the accuracy of {\em Binoculars}.
The biggest impact we observe arises when asking for pirate-sounding output, and this only decreases the sensitivity (increases the false negative rate) by $1\%$; see Figure~\ref{fig:m4-prompting-strategy}. Table~\ref{tab:orca-samples} records generations based on a sample prompt employing the specified prompting strategies.


Next, we also want to test whether arbitrary mistakes, hashcodes, or other kinds of random (and random seeming) strings bias our model towards false positives.  
To test the impact of randomness, we generate random sequences of tokens from the Falcon tokenizer, and score them with \textit{Binoculars} as usual. 
We plot a histogram of this distribution in Figure~\ref{fig:gibberish-text}.
We find that {\em Binoculars} confidently scores this as human, with a mean score around $1.35$ for Falcon (humans have a mean of around $1$).
This is expected, as trained LLMs are strong models of language and exceedingly unlikely to ever generate these completely random sequences. In particular, the generation of these random sequences is even less likely than the generation of perfect human-written text by chance. 

\section{Discussion and Limitations}
\label{sec:discussion}

We present {\em Binoculars}, a method for detecting LLM output in the zero-shot case in which no data is available from the generation model. 
We speculate that this transferability arises from the similarity between modern LLMs, as they all use nearly identical transformer components and are likely trained in large part on Common Crawl (\href{https://commoncrawl.org}{commoncrawl.org}) data from similar time periods.  
As the number of open source LLMs rapidly increases, the ability to detect multiple LLMs with a single detector is a major advantage of {\em Binoculars}, for example when used for platform moderation. 
Our study has a number of limitations.
Due to limited GPU memory, we do not perform broader studies with larger (30B+) open-source models. 
Further, we focus on the problem setting of detecting machine-generated text in normal use, and we do not consider explicit efforts to bypass detection.
Finally, there are other non-conversational text domains, such as source code, which we do not investigate in this study. 



\section*{Impact Statement} 
\label{sec:broader-impact}

Language model detection may be a key technology to reduce harm, whether to monitor machine-generated text on internet platforms and social media, filter training data, or identify responses in chat applications. 
Nevertheless, care has to be taken so that detection mechanisms actually reduce harm, instead of proliferating or increasing it. 
We provide an extensive reliability investigation of the proposed \textit{Binoculars} mechanisms in Section~\ref{sec:edge-cases}, and believe that this is a significant step forward in terms of reliability, for example when considering domains such as text written by non-native speakers. 
Yet, we note that this analysis is only a first step in the process of deploying LLM detection strategies and does not absolve developers of such applications from carefully verifying the impact on their systems. 
We especially caution that the existence of LLM detectors does not imply that using them is worthwhile in all scenarios. 
 
Also, we explicitly highlight that we consider the task of detecting ``naturally'' occurring machine-generated text, as generated by LLMs in common use cases. 
We understand that no detector is perfect and we do not guarantee any performance in settings where a motivated adversary tries to fool our system. We also note that our detector does not provide an explanation or interpretation of its predictions for any given sample and thus is black-box in nature.
We present a thorough evaluation across a wide variety of test sources, but we maintain that directed attempts to bypass our classifier might be possible, as is often the case for systems that rely on neural networks. 

\section*{Acknowledgments}
This work was made possible by the ONR MURI program and the AFOSR MURI program. Commercial support was provided by Capital One Bank, the Amazon Research Award program, and Open Philanthropy. Further support was provided by the National Science Foundation (IIS-2212182), and by the NSF TRAILS Institute (2229885). Jonas Geiping was supported by the Tübingen AI Center.

\bibliography{main}
\bibliographystyle{icml2024}

\appendix
\onecolumn
\section{Appendix}

\label{sec:appendix-experiments}
\subsection{Experimental Details}

\subsubsection{Dataset Generation}
\label{sec:appendix-experiments-data-gen}

Using human-written samples from CC News, CNN and Pubmed datasets, we prompt LLaMA-2-13B and Falcon-7B to generate corresponding machine text. To do so, we peel off the first 50 tokens of each human sample and use it as a prompt to generate up to 512 tokens of machine output.
We then remove the human prompt from the generation and only use the purely machine-generated text in our machine-text datasets. 

\subsubsection{Out of Domain Threshold Tuning}
\label{sec:appendix-threshold}

\begin{figure}[ht!]
    \centering
    \includegraphics[width=0.75\textwidth]{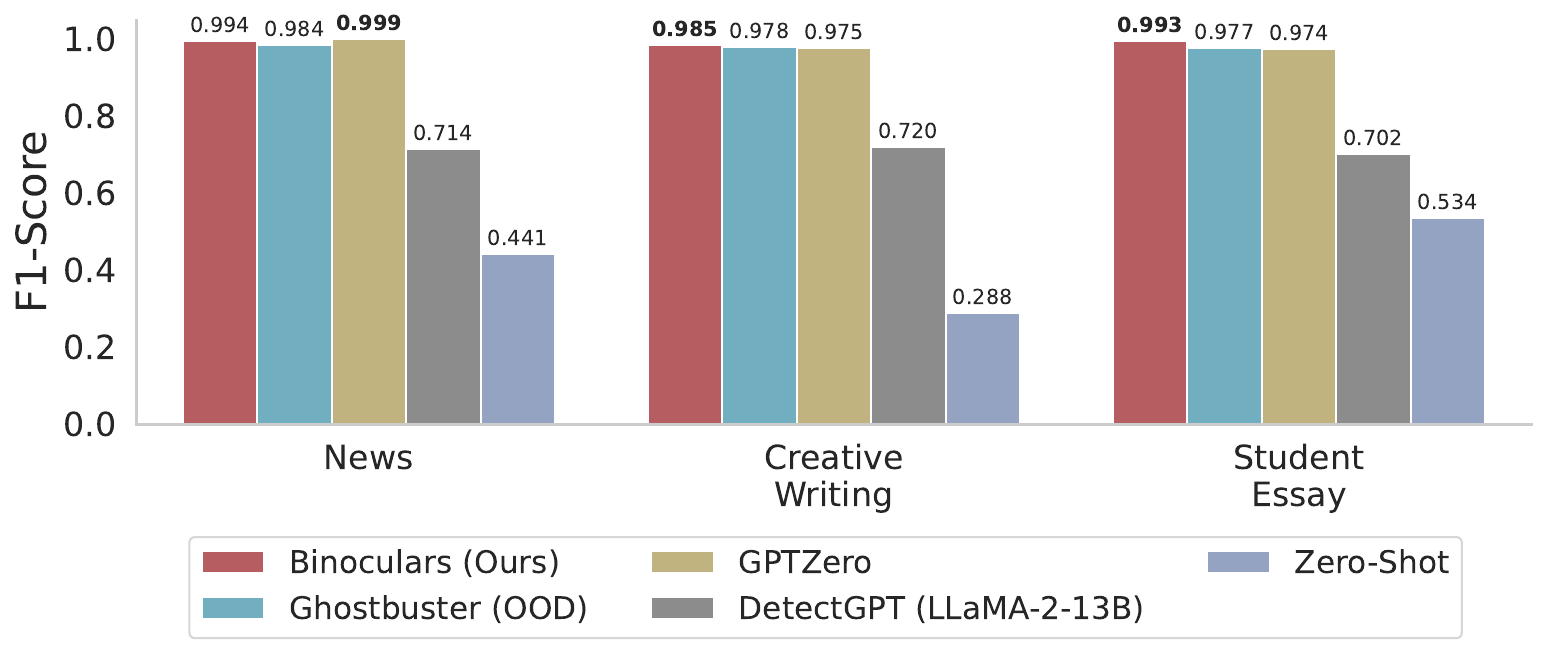}
    \caption{F1 scores for detection of ChatGPT-generated text indicate that several detectors perform similarly. We discuss below how this metric can be a poor indicator of performance at low FPR.}
    \label{fig:gh-dataset-performance}
\end{figure}

As motivated in section \ref{sec:metrics}, we evaluate detectors True-Positive-Rate while operating under ultra-low False-Positive Rate (TRP$@$ 0.01\%FPR). This metric, like AUC, is threshold-agnostic that captures the discrimination power at the desired low False-Positive Rate of the respective method.  When presenting F1-Score (fig. \ref{fig:gh-dataset-performance}), Recall and Precision (fig. \ref{fig:m4-domain-with-baselines}, \ref{fig:m4}, \ref{fig:multi-lingual}), and False Negative Rate (fig. \ref{fig:m4-prompting-strategy}) we use a purely ``out-of-domain'' tuned threshold to separate machine and human text. 


We set the threshold using the combination of training splits from all of our reference datasets: News, Creative Writing, and Student Essay datasets from \citet{verma_ghostbuster_2023}, which are generated using ChatGPT.  We also compare detectors on LLaMA-2-13B and Falcon-7B generated text with prompts from CC News, CNN, and PubMed datasets. All of these datasets have an equal number of human and machine-generated text samples. We optimize using accuracy and fix our threshold globally using these datasets. For all datasets, we use prefix of 512 tokens for each document, unless explicitly mentioned otherwise. 

In order to meet the ``out-of-domain'' claim when evaluating News, Creative Writing, and Student Essay datasets by Ghostbuster paper \citep{verma_ghostbuster_2023} we do not include them in the threshold determination and only use tune threshold on from CC News, CNN, and PubMed (generated via LLaMA and Falcon).

\subsubsection{Baseline Details}
\label{baselines-impl-deets}
As described in the section \ref{sec:intro}, we choose baselines with emphasis on post-hoc, out-of-domain (zero-shot), and black-box detection scenarios. These open source detector Ghostbuster \citep{verma_ghostbuster_2023}, the commercially deployed GPTZero\footnote{\href{https://gptzero.me/}{https://gptzero.me/}}, DetectGPT \citep{mitchell_detectgpt_2023} and its efficient version Fast-DetectGPT \citep{fastdetectgpt} and DNA-GPT \citep{dnagpt2023} to compare detection performance across various datasets in Section~\ref{sec:results}. 

We use out-of-domain version of all of these baselines (applicable only for Ghostbuster) for a fair comparison with our method. If a threshold is provided with original work, we use it for hard prediction otherwise we optimize threshold identical to our method on same datasets for fair and identical comparison.

For DetectGPT, we use LLaMA-2-13B for scoring and T5 model \citep{raffel2023t5} for mask filling for all datasets (even ones generated using LLaMA-2-13B). For Fast-DetectGPT, as described in original work, we use GPT-J-6B and GPT-Neo-2.7B for reference and scoring models respectively for all datasets. We use \textit{gpt-3.5-turbo-instruct} API (March 2024) for suffix prediction in DNA-GPT implementation. GPTZero, a closed sourced API, was queried in September 2023 for our experiments.

\label{sec:app}

\subsection{Benchmark Performance}
\label{f1-score-appendix}

\textbf{ChatGPT Text Detection.} F1-scores on ChatGPT dataset released by \citep{verma_ghostbuster_2023}. The numbers for Zero-Shot baseline method are taken from the same work.  

\subsection{Ablation Studies}
\label{sec:ablations}

\textbf{Comparison to Other Scoring Model Pairs.} 

\begin{table}[htbp]
    \centering
    \footnotesize
    \caption{Other combinations of scoring models, evaluated on our reference datasets as described in the main body.}
    \hspace*{-1cm}
    \begin{tabular}{llp{1.5cm}p{1.5cm}p{1.5cm}p{1.5cm}}
        \toprule
          PPL Scorer ($\mathcal{M}_1$) 
        & X-Cross PPL Scorers ($\mathcal{M}_1'$, $\mathcal{M}_2$)
        & TPR at $0.01\%$ FPR
        & \update{TPR at $0.1\%$ FPR}
        & \update{F1-Score}
        & \update{AUC}
        \\
        \midrule
        Falcon-7B-Instruct     & Falcon-7B, Falcon-7B-Instruct     & 100.0000      & 100.0000    & 1.0000         & 1.0000   \\
        Llama-2-13B            & Llama-13B, Llama-2-13B            & 99.6539       & 99.6539     & 0.9982         & 0.9999   \\
        Llama-2-7B             & Llama-7B, Llama-2-7B              & 99.3079       & 99.3079     & 0.9965         & 0.9998   \\
        Llama-2-13B            & Llama-13B, Llama-2-13B            & 98.3549       & 98.3549     & 0.9913         & 0.9997   \\
        Falcon-7B-Instruct     & Falcon-7B, Falcon-7B-Instruct     & 98.7200       & 99.1600     & 0.9953         & 0.9996   \\
        Falcon-7B-Instruct     & Falcon-7B, Falcon-7B-Instruct     & 94.9200       & 99.4000     & 0.9963         & 0.9996   \\
        Llama-2-7B             & Llama-7B, Llama-2-7B              & 95.8441       & 97.5757     & 0.9922         & 0.9996   \\
        Llama-2-13B            & Llama-13B, Llama-2-13B            & 98.6400       & 99.0400     & 0.9953         & 0.9995   \\
        Llama-2-7B             & Llama-7B, Llama-2-7B              & 98.8000       & 99.2800     & 0.9959         & 0.9995   \\
        Llama-2-7B             & Llama-7B, Llama-2-7B              & 98.1600       & 98.6000     & 0.9937         & 0.9992   \\
        Llama-2-13B            & Llama-13B, Llama-2-13B            & 98.4000       & 98.7200     & 0.9943         & 0.9992   \\
        Falcon-7B-Instruct     & Falcon-7B, Falcon-7B-Instruct     & 94.1125       & 97.9220     & 0.9926         & 0.9992   \\
        Falcon-7B-Instruct     & Falcon-7B, Falcon-7B-Instruct     & 93.5000       & 93.5000     & 0.9875         & 0.9990   \\
        Falcon-7B-Instruct     & Falcon-7B, Falcon-7B-Instruct     & 92.0000       & 92.0000     & 0.9918         & 0.9990   \\
        Llama-2-7B             & Llama-7B, Llama-2-7B              & 94.0000       & 94.0000     & 0.9850         & 0.9989   \\
        Llama-2-7B             & Llama-7B, Llama-2-7B              & 98.0000       & 98.0000     & 0.9956         & 0.9988   \\
        Falcon-7B-Instruct     & Falcon-7B, Falcon-7B-Instruct     & 72.6957       & 72.7857     & 0.9908         & 0.9988   \\
        Llama-2-13B            & Llama-13B, Llama-2-13B            & 97.8750       & 97.8750     & 0.9931         & 0.9987   \\
        Llama-2-13B-Chat       & Llama-2-13B, Llama-2-13B-Chat     & 71.3199       & 82.6799     & 0.9846         & 0.9986   \\
        Llama-2-13B            & Llama-13B, Llama-2-13B            & 97.5000       & 97.5000     & 0.9875         & 0.9985   \\
        Falcon-7B-Instruct     & Falcon-7B, Falcon-7B-Instruct     & 97.5778       & 97.5778     & 0.9930         & 0.9983   \\
        Falcon-7B-Instruct     & Falcon-7B, Falcon-7B-Instruct     & 23.3076       & 48.3732     & 0.9842         & 0.9975   \\
        Llama-2-13B            & Llama-13B, Llama-2-13B            & 0.3200        & 32.0800     & 0.9840         & 0.9968   \\
        Llama-2-13B-Chat       & Llama-2-13B, Llama-2-13B-Chat     & 20.9172       & 60.0671     & 0.9763         & 0.9968   \\
        Llama-2-13B            & Llama-13B, Llama-2-13B            & 47.1476       & 69.2953     & 0.9747         & 0.9964   \\   
\bottomrule
\label{ablation:other-scorers}
\end{tabular}
\end{table}

\begin{figure}
    \centering
    \includegraphics[trim=1cm 10cm 6cm 1cm,width=\textwidth,clip]{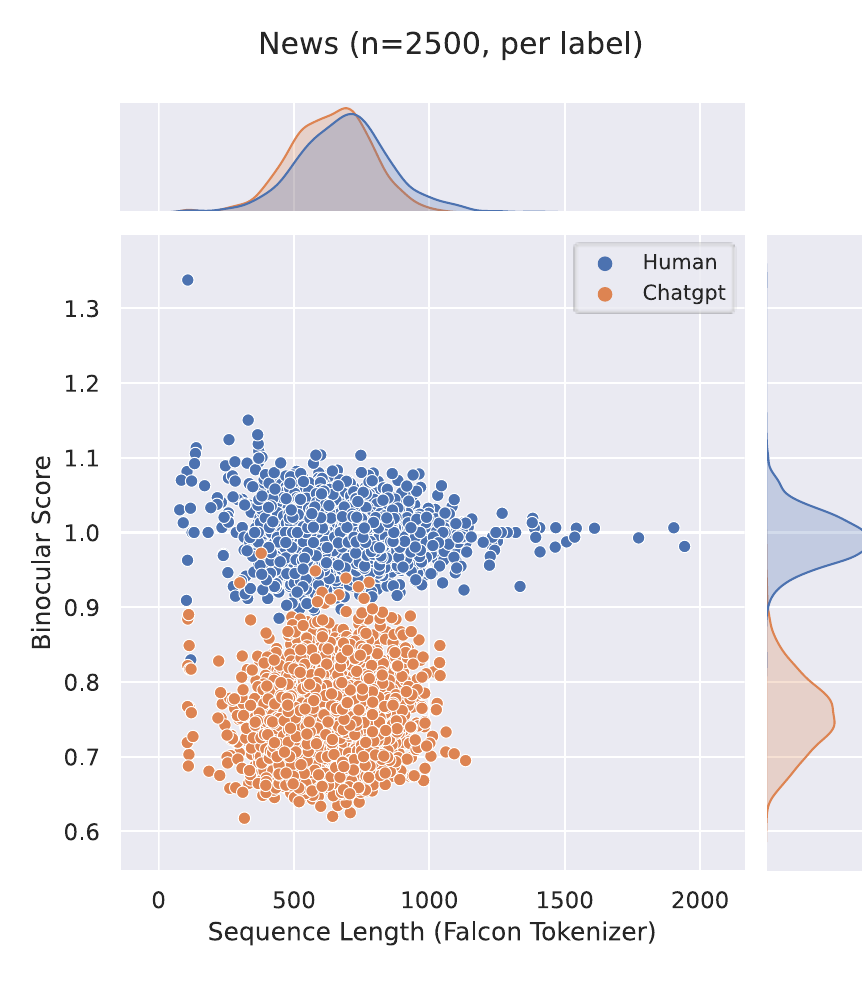}
    \caption{A closer look at the actual distribution of scores in terms of sequence length for the Ghostbuster news dataset.}
    \label{fig:double-scatter}
\end{figure}

\textbf{String Length.}
Is there a correlation between \emph{Binoculars} score and sequence length? Such correlations may create a bias towards incorrect results for certain lengths.
In Figure~\ref{fig:double-scatter}, we show the joint distribution of token sequence length and \emph{Binocular} score. Sequence length offers little information about class membership.

\begin{figure*}[t!]
    \centering
    \includegraphics[width=0.5\textwidth]{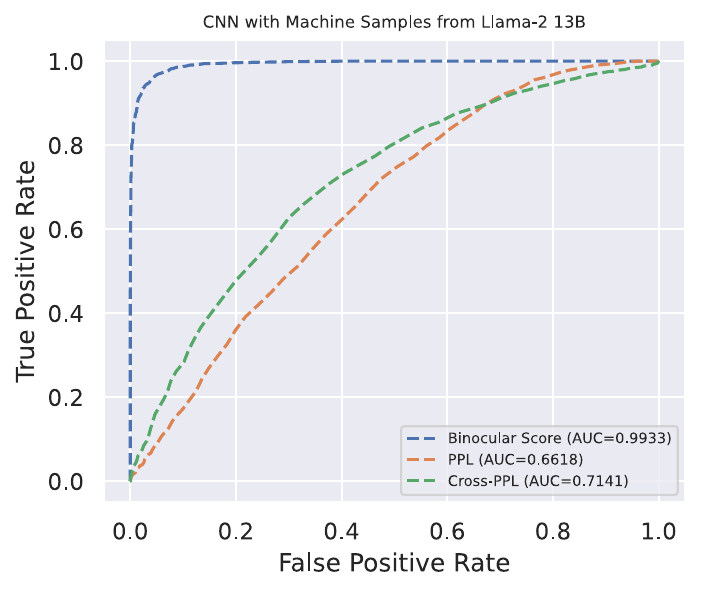}
    \caption{Perplexity and Cross-perplexity are not strong detectors on their own.}
    \label{fig:ablation-ppl-xppl}
\end{figure*}

\textbf{Score Components.} 
Perplexity is used by many detecting formulations in isolation. We show in Figure~\ref{fig:ablation-ppl-xppl} that both perplexity and cross-perplexity are not effective detectors in isolation. Table~\ref{tab:ppl-v-xppl-results} show the results where we compute PPL and X-PPL with different model families viz. LLaMA-2 and Falcon. 

\begin{table}[ht!]
\caption{Over various datasets, we show that perplexity alone or cross-perplexity alone are poor predictors of human versus machine, whereas Binoculars perform well even at low false-positive rates (FPR).}
\label{tab:ppl-v-xppl-results}
\footnotesize
\centering
  \begin{tabular}{llcccccc}
    \toprule
                 & & & \multicolumn{4}{c}{True Positive Rate} \\
        Dataset  &           Detector &            AUC &    @ 0.01\% FPR &     @ 0.1\% FPR &      @ 1\% FPR &      @ 5\% FPR \\ 
    \midrule
                 &         Falcon PPL &           1.00 &            0.86 &            0.86 &           0.94 &           0.98 \\ 
                 &       Falcon X-PPL &           0.94 &            0.56 &            0.56 &           0.59 &           0.79 \\ 
        Writing  &          LLaMA PPL &           0.99 &            0.86 &            0.86 &           0.92 &           0.98 \\ 
         Prompts &        LLaMA X-PPL &           0.86 &            0.04 &            0.04 &           0.10 &           0.43 \\ 
                 &  Binoculars-Falcon &  \textbf{1.00} &            0.93 &            0.93 &           0.96 &  \textbf{1.00} \\ 
                 &   Binoculars-LLaMA &  \textbf{1.00} &   \textbf{0.95} &   \textbf{0.95} &  \textbf{0.98} &  \textbf{1.00} \\ 
    \midrule
                 &         Falcon PPL &           0.99 &            0.65 &            0.77 &           0.90 &           0.95 \\ 
                 &       Falcon X-PPL &           0.85 &            0.04 &            0.12 &           0.29 &           0.53 \\ 
         News    &          LLaMA PPL &           0.98 &            0.67 &            0.71 &           0.89 &           0.95 \\ 
                 &        LLaMA X-PPL &           0.26 &            0.00 &            0.00 &           0.00 &           0.01 \\ 
                 &  Binoculars-Falcon &  \textbf{1.00} &            0.95 &   \textbf{0.99} &  \textbf{1.00} &  \textbf{1.00} \\ 
                 &   Binoculars-LLaMA &  \textbf{1.00} &   \textbf{0.99} &   \textbf{0.99} &  \textbf{1.00} &  \textbf{1.00} \\ 
    \midrule
                 &         Falcon PPL &           1.00 &            0.78 &            0.78 &           0.88 &           0.99 \\ 
                 &       Falcon X-PPL &           0.93 &            0.25 &            0.25 &           0.38 &           0.70 \\ 
        Essay    &          LLaMA PPL &           0.99 &            0.42 &            0.42 &           0.90 &           0.98 \\ 
                 &        LLaMA X-PPL &           0.80 &            0.01 &            0.01 &           0.04 &           0.16 \\ 
                 &  Binoculars-Falcon &  \textbf{1.00} &            0.98 &            0.98 &           0.99 &  \textbf{1.00} \\ 
                 &   Binoculars-LLaMA &  \textbf{1.00} &   \textbf{0.99} &   \textbf{0.99} &  \textbf{1.00} &  \textbf{1.00} \\ 
    \bottomrule
  \end{tabular}
\end{table}

\begin{figure}[ht!]
    \centering\includegraphics[width=0.6\textwidth]{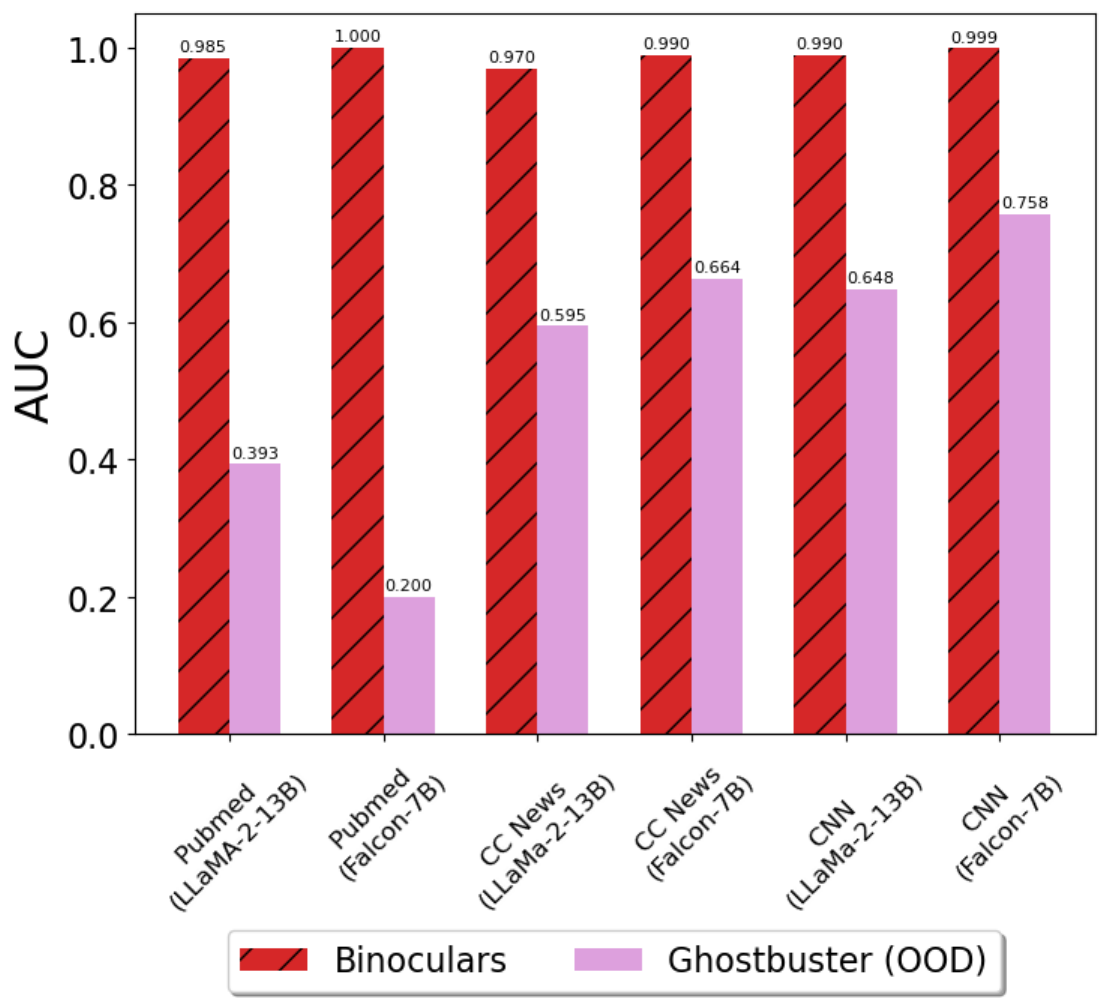}
    \caption{Comparison of Ghostbuster and Binoculars AUC on PubMed, CCNews and CNN datasets. 
    }
    \label{fig:performance_our_datasets}
\end{figure}

\subsection{Other famous texts}
\label{sec:app-famous-text}

Two songs by Bob Dylan further demonstrate this behavior.
\emph{Blowin' In The Wind}, a famous Dylan track has a much lower Falcon perplexity than his unreleased song \emph{To Fall In Love With You} (logPPL values are 1.11 and 3.30, respectively.)
It might be reasonable for famous songs to get classified as machine text and they are more likely output than less famous songs. 
\emph{Binoculars}, however, labels both of these samples confidently as human samples (with scores of 0.92, and 1.01, respectively).

\begin{table}[!ht]
    \centering
    \caption{Case Studies of Text Samples likely to be memorized by LLMs. \label{tab:memorization_studies}}
    \footnotesize
    \hspace*{-1cm}
    \begin{tabular}{lm{1in}m{1in}m{1in}m{0.5in}}
    \toprule
        Human Sample & PPL (Falcon 7B Instruct) & Cross PPL (Falcon 7B, Falcon 7B Instruct) & Binoculars Score & Predicted as Human-Written\\ 
        \midrule
        US Constitution & 0.6680 & 0.8789 & 0.7600 & \textcolor{red}{\faTimes{}} \\ 
        ``I have a dream speech" & 1.0000 & 1.2344 & 0.8101 & \textcolor{red}{\faTimes{}} \\ 
        Snippet from Cosmos series & 2.3906 & 2.8281 & 0.8453 & \textcolor{red}{\faTimes{}} \\ 
        Blowin' In the Wind (song) & 1.1172 & 1.2188 & 0.9167 & \textcolor{green}{\faCheck{}}\\ 
        Oscar Wilde's quote & 2.9219 & 3.0781 & 0.9492 & \textcolor{green}{\faCheck{}}\\ 
        Snippet from White Night & 2.6875 & 2.8125 & 0.9556 & \textcolor{green}{\faCheck{}} \\ 
        Wish You Were Here & 2.5000 & 2.5938 & 0.9639 & \textcolor{green}{\faCheck{}} \\ 
        Snippet from Harry Potter book & 2.5938 & 2.6875 & 0.9651 & \textcolor{green}{\faCheck{}} \\ 
        First chapter of A Tale of Two Cities & 2.7188 & 2.7500 & 0.9886 & \textcolor{green}{\faCheck{}} \\ 
        Snippet from Crime and Punishment & 2.8750 & 2.9063 & 0.9892 & \textcolor{green}{\faCheck{}} \\ 
        To Fall In Love With You (song) & 3.2969 & 3.2656 & 1.0096 & \textcolor{green}{\faCheck{}}\\ 
        \bottomrule
    \end{tabular}
    \label{famous-quotes}
\end{table}

\subsection{Identical Scoring Model}
\label{sec:m1-m2-identical}

\begin{figure}[ht!]
    \centering
    \includegraphics[width=1\textwidth]{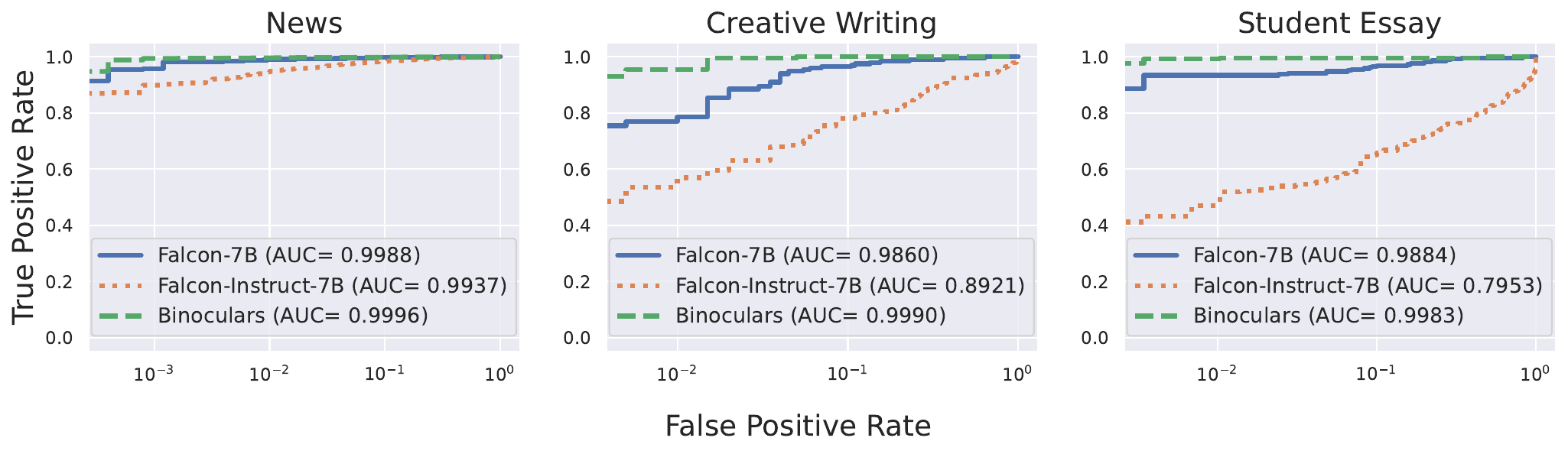}
    \caption{\update{\textbf{AUC Curve} Binoculars score using identical $\mathcal{M}_1$ and $\mathcal{M}_2$ models using Falcon-7B and Falcon-7B-Instruct. 
    }}
    \label{fig:m1-m2-same}
\end{figure}

\update{We inspect Binocular's performance when we choose to use identical $\mathcal{M}_1$ and $\mathcal{M}_1$ models in equation (\ref{eq:binoculars}). We use Falcon-7B and Falcon-7B-Instruct models and compare the two performances with Binoculars Score over dataset by \citep{verma_ghostbuster_2023} in Figure~\ref{fig:m1-m2-same}. We observe although the vanilla Binoculars score is best over 3 domains, using Falcon-7B as input models is competitive. }

\begin{table}[ht!]
\begin{center}
\caption{ Instructions appended in system prompts for 3 different strategies.}
\label{tab:modified-prompts-system-prompts}
\begin{tabular}{  >{\arraybackslash}p{0.22\linewidth}  m{4in}  } 
 \toprule
 Prompting Strategy & Instruction appended to the default system prompt \\
 \midrule
 Carl Sagan & \footnotesize{Write in the voice of Carl Sagan.} \\ 
 \\
 Non-Robotic & \footnotesize{Write your response in a way that doesn’t sound pretentious or overly formal. Don’t use robotic-sounding words like `logical’ and `execute.’  Write in the casual style of a normal person.} \\ 
 \\
 Pirate & \footnotesize{Write in the voice of a pirate.} \\
 \bottomrule
\end{tabular}
\label{tab:orca-system-prompts}
\end{center}
\end{table}

\subsection{Modified System Prompts}
\label{sec:app-modified-prompts}

We test Binoculars' and comparable baselines' performances in Section~\ref{sec:modified-prompts}  on multiple prompting strategies. We prompt LLaMA-2-13B-chat with samples from the Open-Orca dataset. In addition to the default sample-specific prompt, we use 3 different versions in which we append instructions into the system prompt. These include instruction to write in the style of Carl Sagan, in a non-robotic tone, and like a pirate. In Table~\ref{tab:modified-prompts-system-prompts} we mention the exact instruction appended to the default system prompts.

\subsection{Random Tokens}
\label{sec:random-data}

\begin{figure}[ht!]
    \centering
    \includegraphics[width=0.5\columnwidth]{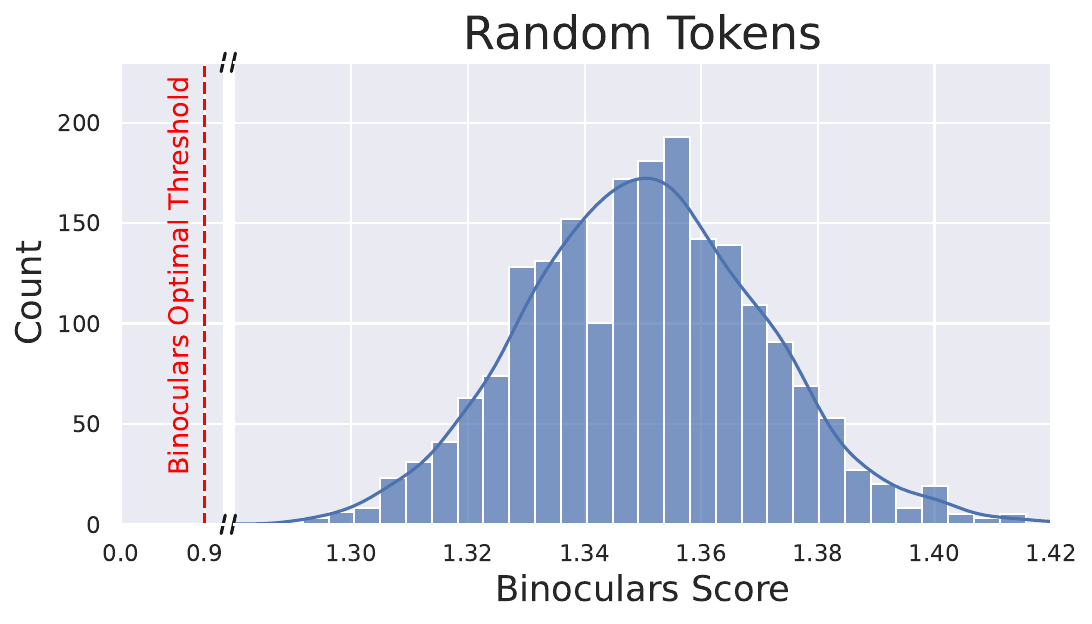}
    
    \caption{Random token sequences fall strictly on the human side of the Binoculars threshold.} 
    \label{fig:gibberish-text}
\end{figure}

Next, we also want to test whether arbitrary mistakes, hashcodes, or other kinds of random (and random-seeming) strings bias our model toward false positives.  
To test the impact of randomness, we generate random sequences of tokens from the Falcon tokenizer and score them with \textit{Binoculars} as usual. 
We plot a histogram of this distribution in Figure~\ref{fig:gibberish-text}.
We find that {\em Binoculars} confidently scores this as human, with a mean score around $1.35$ for Falcon (humans have a mean of around $1$).
This is expected, as trained LLMs are strong models of language and exceedingly unlikely to ever generate these completely random sequences of tokens in any situation. In particular, the generation of these random sequences is even less likely than the generation of perfect human-written text by chance. 

\begin{figure}[ht!]
    \centering
    \includegraphics[width=0.5\columnwidth]{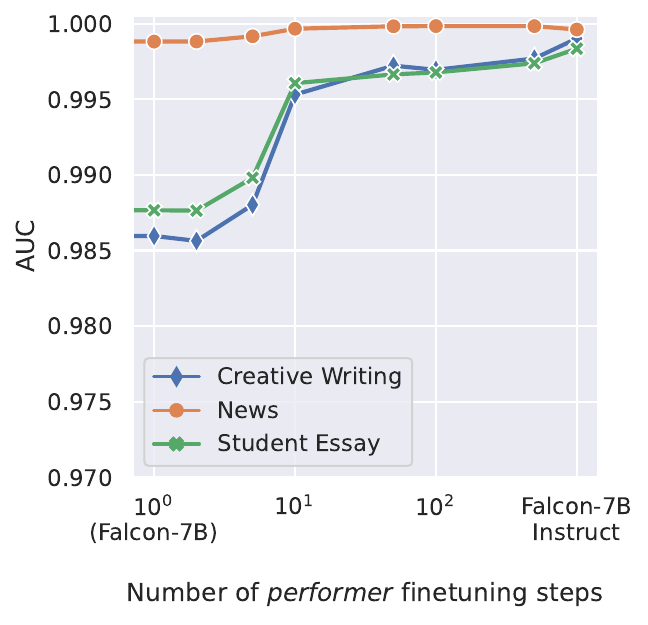}
    
    \caption{We Find that fully finetuned $\mathcal{M}_2$ (i.e., Falcon-7B-Instruct) achieves the best performance, while fine-tuning on the instruction dataset (Alpaca) further enhances performance. This experiment complements the findings from Figure~\ref{fig:m1-m2-same}.} 
    \label{fig:m2-finetuning}
\end{figure}

\subsection{Performer Model Ablation}
In this experiment, we aimed to determine the highest-performing model (M2) to use in the Binoculars setup. In Appendix~\ref{sec:m1-m2-identical}, we demonstrated that using identical scoring models (Falcon-7B) is not the optimal choice, and Falcon-7B-instruct as performer model yields a better detector. Therefore, we know that instruction tuning does help improve Binoculars' performance. We fine-tuned Falcon-7B on the Alpaca instruction tuning dataset and used it as the performer model. We benchmark Binoculars' performance with performers trained at various steps: 0 (identical model case), 1, 10, 50, 100, 500, and finally using Falcon-7B-instruct (original formulation). We observed that the fully fine-tuned performer (Falcon-7B-Instruct) achieved the best detection performance, and this performance increase was nearly monotonic with instruction fine-tuning.

\textbf{Experiment Details.} We finetune pretrained Falcon-7B on alpaca instruction dataset with 5e-5 learning rate and 65K tokens batch size (32 samples * 2048 block size) with cosine annealing ratio of 3\% on 4 A5000 GPUs using FSDP distributed training.

\subsection{Confidence estimates for Binoculars Peformance}

We report the standard error on our reported AUC and TPR @ 0.01\% FPR in Figure 1 to provide confidence estimates around these figures. We achieve this by creating 20 one-third-sized subsamples from the original set using stratified bootstrapping, ensuring a 50-50 class mix.

\begin{table}[ht!]
\caption{Standard error for reported metrics from the main paper.}
\centering
\begin{tabular}{llll}
\hline
Dataset Name & AUC     & TPR @ 0.01\% FPR &  \\
\hline
News             & 2.12e-5 & 2.92e-3          &  \\
Creative Writing & 2.50e-4 & 2.15e-3          &  \\
Student Essay    & 8.99e-5 & 3.89e-3          &  \\
\hline
\end{tabular}
\end{table}

\subsection{Binoculars Peformance on GPT4 and Gemini-Pro}

We evaluate Binoculars' performance on state-of-the-art APIs as of March 2024. We randomly sample instruction and system prompt pairs from the Open Orca dataset and use the GPT-4 and Gemini APIs to generate text. We observe a very low false negative rate for Gemini, while it is considerably high for GPT-4.

\begin{table}[!ht]
\caption{False negative rate on samples generated by state-of-the-art generation API}
\vspace{2pt}
\centering
\footnotesize
\begin{tabular}{llccccc}
\toprule
    API Name & Version & \# Correct& \# Total & Source & Acc. & False Negative Rate \\
    \midrule
    gemini-1.0-pro-latest & March 2024 & 125 & 129 & Open Orca & 96.89\% & 3.10\% 
    \\
    gpt-4 & March 2024 & 54 & 129 & Open Orca & 41.86\% & 58.13\% \\ \bottomrule
\end{tabular}
\end{table}

\end{document}